\documentclass{article}

\usepackage{microtype}
\usepackage{graphicx}
\usepackage{subfigure}
\usepackage{booktabs} % for professional tables
\usepackage{multirow}
\usepackage{hyperref}

\usepackage[accepted]{icml2024}

\usepackage{amsmath}
\usepackage{amssymb}
\usepackage{mathtools}
\usepackage{amsthm}

\usepackage[capitalize,noabbrev]{cleveref}

\theoremstyle{plain}

\theoremstyle{definition}

\theoremstyle{remark}

\usepackage[textsize=tiny]{todonotes}

\icmltitlerunning{Submission and Formatting Instructions for ICML 2024}

\begin{document}

\twocolumn[
\icmltitle{HGCN2SP: Hierarchical Graph Convolutional Network for Two-Stage Stochastic Programming}



\icmlsetsymbol{equal}{*}

\begin{icmlauthorlist}
\icmlauthor{Yang Wu}{iacas,ucas}
\icmlauthor{Yifan Zhang}{iacas,nj,air}
\icmlauthor{Zhenxing Liang}{iacas,ucas}
\icmlauthor{Jian Cheng}{iacas,nj,air}
\end{icmlauthorlist}

\icmlaffiliation{iacas}{Institute of Automation, Chinese Academy of Sciences, Beijing, China}
\icmlaffiliation{ucas}{School of Artificial Intelligence, University of Chinese Academy of Sciences, Beijing, China}
\icmlaffiliation{nj}{University of Chinese Academy of Sciences, Nanjing, Nanjing, China}
\icmlaffiliation{air}{AIRIA, Nanjing, China}

\icmlcorrespondingauthor{Yifan Zhang}{yfzhang@nlpr.ia.ac.cn}

\icmlkeywords{Machine Learning, ICML}

\vskip 0.3in
]



\printAffiliationsAndNotice{\icmlEqualContribution} % otherwise use the standard text.

\begin{abstract}
Two-stage Stochastic Programming (2SP) is a standard framework for modeling decision-making problems under uncertainty. While numerous methods exist, solving such problems with many scenarios remains challenging. Selecting representative scenarios is a practical method for accelerating solutions. However, current approaches typically rely on clustering or Monte Carlo sampling, failing to integrate scenario information deeply and overlooking the significant impact of the scenario order on solving time. To address these issues, we develop HGCN2SP, a novel model with a hierarchical graph designed for 2SP problems, encoding each scenario and modeling their relationships hierarchically. The model is trained in a reinforcement learning paradigm to utilize the feedback of the solver. The policy network is equipped with a hierarchical graph convolutional network for feature encoding and an attention-based decoder for scenario selection in proper order. Evaluation of two classic 2SP problems demonstrates that HGCN2SP provides high-quality decisions in a short computational time. Furthermore,  HGCN2SP exhibits remarkable generalization capabilities in handling large-scale instances, even with a substantial number of variables or scenarios that were unseen during the training phase.
\end{abstract}

\section{Introduction}
\label{label:intro}
Two-stage Stochastic Programming (2SP) is a powerful framework for modeling decision-making under uncertainty. It starts with the first stage decision, followed by the determination of specific parameters (a realization of certain \textit{scenario}). Then, the second-stage decisions are made based on these parameters. The aim is to minimize the total cost of the first-stage decisions and the expected cost of potential scenarios, ensuring optimal first-stage decision-making.
\par Consider the occurrence of a disaster as an uncertain event. In the first stage, decisions on the location and quantity of emergency supplies need to be made without knowing the number of victims or the extent of the damage. The goal is to make the optimal first-stage decisions that can swiftly address the needs of disaster relief, regardless of specific disaster details, like destroyed roads. This approach applies not only to disaster management \cite{noyan2012risk, grass2016two} but to fields including network design, distributed energy systems, facility location, and blood supply chain management \cite{santoso2005stochastic, zhou2013two, bieniek2015note, dillon2017two, abbasi2020predicting}.

Within this specialized field, various traditional algorithms have been developed.  Among them,  the L-shaped method is particularly notable, though it tends to converge slowly with numerous scenarios \cite{birge2023convergence} and struggles when second-stage decisions involve integers \cite{sen2001stochastic}.  Sample Average Approximation (SAA) is another widely used method that approximates the problem by sampling scenarios. However, this method faces a balance challenge, where few samples result in inaccuracies, and too many complicate the solution. Scenario reduction is a critical technique extensively explored in literature \cite{ dupavcova2003scenario, heitsch2003scenario, morales2009scenario, dvorkin2014comparison, rujeerapaiboon2022scenario, bertsimas2023optimization}. It selects a representative subset of scenarios to accelerate the solution process. Despite the significance,  existing scenario reduction methods often overlook problem-specific details or are limited to theoretical results, lacking practical algorithms \cite{keutchayan2023problem}.

\begin{figure*}[ht]
\vskip 0.2in
\begin{center}
\centerline{\includegraphics[width=0.98\textwidth, trim=0cm 0cm 0cm 0cm, clip]{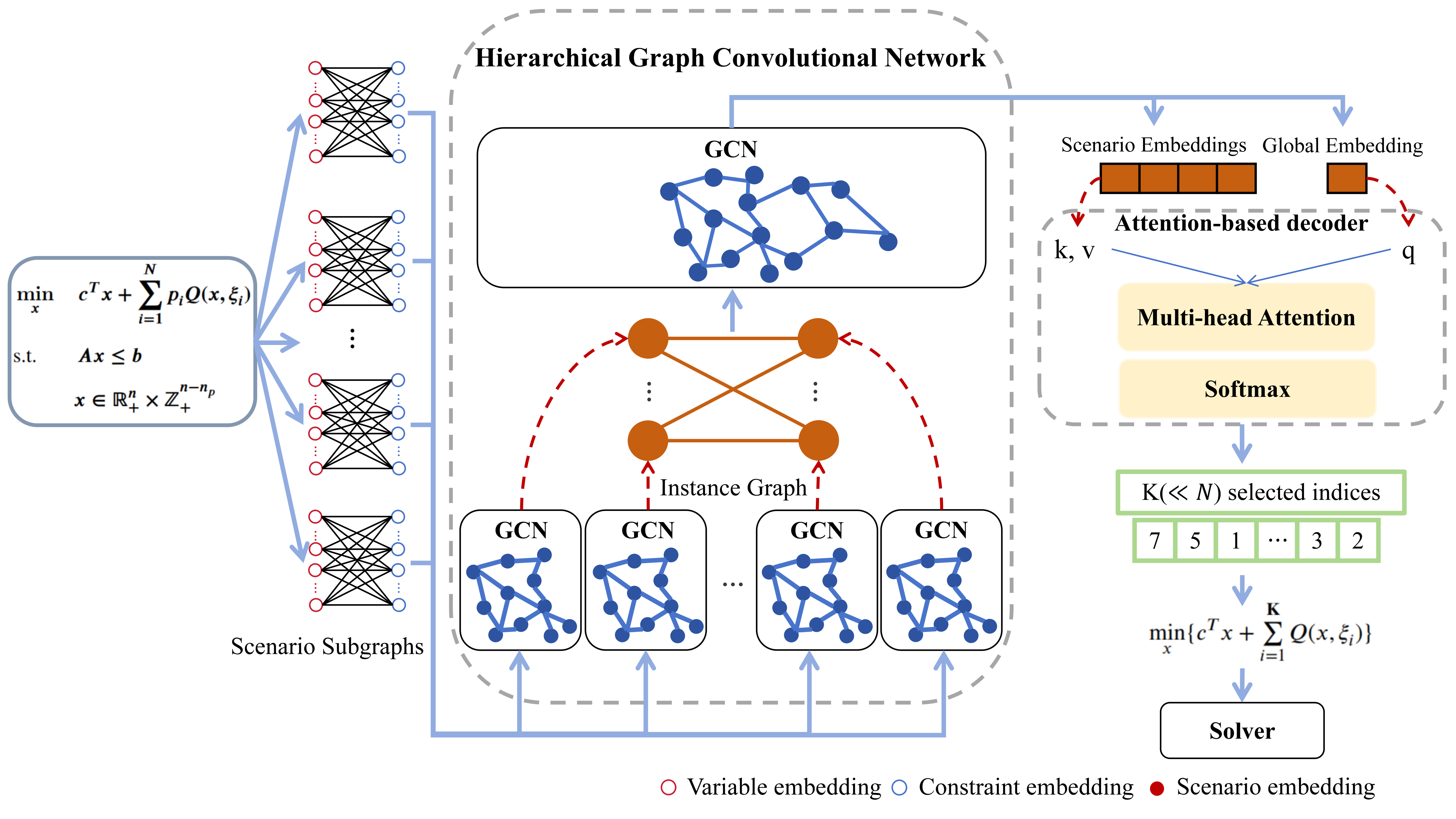}}
\caption{An overview of HGCN2SP. For each 2sp instance, we initially combine individual scenarios with the first stage to form scenario subgraphs. These subgraphs are then input into a hierarchical graph convolutional network (GCN) to derive representations. In this network, each scenario subgraph first obtains embeddings from the low-level GCN, building an instance graph. The high-level GCN then utilizes this graph to generate the final and global embeddings. An attention-based decoder leverages these embeddings to select $k$ scenarios sequentially, which are transformed into an equivalent MIP problem. Ultimately, a solver is employed to derive the solution. }
\label{fig:pipeline}
\end{center}
\vskip -0.2in
\end{figure*}

\par Given the rapid development of Machine Learning (ML), especially in decision-making tasks with perfect information, recent efforts have integrated machine learning methods to address 2SP, with various studies contributing to this field \cite{nair2018learning, larsen2018predicting,bengio2020learning, patel2022neur2sp}. A notable work in this context is \cite{wu2022learning}, which applies self-supervised learning for generating scenario representations and clustering for scenario reduction. However, this approach treats scenarios individually and only considers their interrelationships during clustering, thereby failing to fully exploit the comprehensive information available in the scenario space.

\par Furthermore, previous studies on scenario reduction deem scenario selection as an isolated task, overlooking the significant fact that scenario reduction alone does not guarantee conclusive results. Instead, after selecting a manageable number of scenarios, these methods convert the 2SP into an equivalent deterministic problem, with advanced solvers then used for the final results.  Therefore, utilizing the feedback of the solver will be beneficial for selection. Moreover, it is worth noting that the order of scenarios significantly impacts the processing time of the solver. Specifically, the sequence of scenarios influences the order of constraints and the choice of the initial basis in the simplex method, which would greatly affect solving time  \cite{bixby1992implementing, li2022learning}. Despite this, the majority of existing scenario reduction methods, dependent on clustering for selection, fail to exploit the advantages offered by sequential information.

To tackle the outlined challenges, we propose HGCN2SP, an innovative model leveraging a hierarchical graph specifically designed for the two-stage nature of 2SP. The lower level of this graph is a bipartite graph that merges each scenario's details with the first-stage information, while the higher level identifies correlations between scenarios. HGCN2SP employs a hierarchical graph convolutional network (GCN) for processing. The low-level GCN extracts scenario embeddings from each bipartite graph, and the high-level GCN explores topological relationships within the scenario space. An attention-based decoder then utilizes these deep embeddings to sequentially select scenarios from the candidate pool.  Once the sequence is determined, we convert these scenarios into an equivalent problem for solver analysis. Additionally, we integrate reinforcement learning to optimally utilize solver feedback, focusing on both the results and solution times of scenario sequences. This method enables us to select representative scenarios, thus achieving an exceptionally accurate approximation of the original problem and enhancing the solver's efficiency.

\par  Our main contributions are as follows:\par 
\textbullet  \enspace  Developing a specialized hierarchical graph specifically tailored for the nature of 2SP. Such a graph effectively captures the core aspects of the problem, offers scalability, and is adaptable to various problem types.

\textbullet \enspace   Introducing a model that integrates RL with GCN for selecting representative scenarios in 2SP and leveraging the solver feedback. Our method considers both the performance and efficiency of solving.

\textbullet  \enspace Achieving decision-making results that surpass or be comparable to existing methods, but in less time. Furthermore, our approach demonstrates strong generalization abilities for problems with larger-scale or more scenarios.

\section{Related Work}

\textbf{Solution methods for 2SP} Solving 2SP is a key focus in operations research. The L-shaped method \cite{laporte1993integer, angulo2016improving} divides 2SP into master and sub-problems, solving them iteratively until convergence. However, it can be slow for complex sub-problems and suits only certain structures. Sample Average Approximation \cite{birge2011introduction, shapiro2021lectures} is another primary method, facing a trade-off: smaller samples lead to greater errors, while larger ones take more time. Scenario reduction, introduced by \cite{dupavcova2003scenario}, reduces computational effort by approximating the original distribution with fewer points. Techniques like the Wasserstein distance measure the differences \cite{henrion2009scenario,rujeerapaiboon2022scenario}, but these methods may not fully leverage problem-specific aspects like the objective function.
Lately, problem-oriented scenario reduction methods have gained attention. They use problem-specific information and are theoretically robust \cite{keutchayan2020scenario, henrion2022problem, fairbrother2022problem}, but practical implementation is often limited. \cite{keutchayan2023problem} developed a method that calculates objective values for each scenario and selects representative ones by projecting scenarios into value space. While effective and parallelizable, this approach is still time-intensive.

\textbf{Learning-based methods for 2SP}
The rapid development in machine learning and combinatorial optimization has led to new solving algorithms. \cite{nair2018learning} proposed an RL-based iterative local search method, but it is limited to binary decision-making. \cite{bengio2020learning} uses supervised learning to predict a representative scenario, simplifying the original problem. \cite{larsen2023fast} accelerates the L-shaped method by substituting machine learning predictions for costly computations. \cite{yilmaz2023deep} employs multi-agent reinforcement learning with two agents for different stages but does not fully address key problem aspects like the objective function coefficient. \cite{patel2022neur2sp} predicts the second-stage objective value using a ReLU neural network, converting it into a mixed integer programming problem (MIP) for optimization. However, this method focuses only on uncertain parameters and initial-stage decisions, thus lacking generalizability and cannot be applied across instances without specific training. \cite{wu2022learning} applies conditional variational autoencoder (CVAE) to learn latent scenario representations for scenario reduction. Yet, this modeling method is limited to %graph-based 
graphical structured 2SP problems. Furthermore, it encodes scenarios individually during training and inference, ignoring the deeper connections between them. 
\par Moreover, while these methods use solvers for final solutions, they do not fully utilize solver feedback for optimization. For example, the solver's computation time is as important as the optimal solution in practical application.

\section{Preliminaries}
\label{label:pre}
In this section, we present a comprehensive overview of 2SP and the bipartite graph representation.
\subsection{Two-stage Stochastic Programming}
The definition of two-stage stochastic programming problems encompasses both deterministic and stochastic (uncertain) components, with a common expression being: 
\begin{align}
    v^{*}=&\min\limits_{x} \quad c^T x + \mathbb{E}_{\xi} [Q(x,\xi)] \label{eq:2sp_expected} \\
    & \text{s.t.} \qquad  Ax \leq b,   \quad  x \in \mathbb{R}^{p_1}\times \mathbb{Z}^{n_1-p_1}, \quad  \nonumber\\ 
     \mbox{where} \,\, & Q(x,\xi):= \min\limits_{y}  \{q_{\xi}^{T} y| W_{\xi}y\leq h_{\xi}-T_{\xi}x \}, \label{eq:Q_value} \\
     & \qquad \quad  y \in \mathbb{R}^{p_2} \times \mathbb{Z}^{n_2-p_2},\nonumber
\end{align}
where $x$ and $y$ denote the decision variables of the first and second stages, respectively, $A\in \mathbb{R}^{m_1\times n_1}$, $b\in \mathbb{R}^{m_1}$, and  $c \in \mathbb{R}^{n_1}$ are the parameters of the first stage and provide deterministic information about the problem. For scenario $\xi$, the corresponding uncertainty parameters are $q_{\xi}\in \mathbb{R}^{n_2}$, $W_{\xi} \in \mathbb{R}^{m_2\times n_2}$, $h_{\xi} \in \mathbb{R}^{m_2}$, and $T_{\xi} \in \mathbb{R}^{m_2 \times n_1}$.

\par As indicated in equation (\ref{eq:Q_value}), $Q(x,\xi)$ is the optimal value of a MIP. Given the exponentially vast or even continuous range of $\xi$, calculating the expected value $\mathbb{E}_{\xi} [Q(x,\xi)]$ is infeasible. Consequently,  SAA is usually employed to uniformly sample an empirical distribution $\{\xi_1,\cdots,\xi_N\}$. Such distribution is then utilized to approximate the expected value, typically achieved by converting equation (\ref{eq:2sp_expected}) into its extensive form (EF), as outlined below:
\begin{align}
    &\min\limits_{x} \quad  c^T x + \sum\limits_{i=1}^{N} p_i Q(x,\xi_i) \label{eq:ef_form}  \\
    &\text{s.t.} \qquad  Ax\leq b, \quad x \in \mathbb{R}^{p_1}\times \mathbb{Z}^{n_1-p_1},\nonumber
\end{align}
where $p_i$ is the probability of scenario $\xi_i$ being sampled. 
\par However, increasing the number of sampled scenarios leads to a rise in both variable dimensions and constraint numbers. This increase, coupled with the complexities of solving MIPs, makes solving the problem impractical.

\par \textbf{Scenario reduction} further simplifies equation (\ref{eq:ef_form}) by downsampling. It attempts to select $k$ scenarios $\{\xi_{j_1},\cdots, \xi_{j_k}\}$ ($k\!\ll\! N$) and obtain $\tilde{x}^{*} \! = \!\arg\min\{c^Tx \! +\! \sum\limits_{i=1}^{k} p_i Q(x,\xi_{j_i})\}$. The aim is to minimize the gap between $f(\tilde{x}^{*})\!\coloneqq \!c^T \tilde{x}^{*}\!+\!\mathbb{E}_{\xi}[Q(\tilde{x}^{*}])$ and $v^{*}$ (see \cref{eq:2sp_expected}). Unlike SAA, scenario reduction moves beyond random sampling by leveraging problem-specific information, such as objective values and constraints. This approach can improve both tractability and interpretability \cite{bertsimas2023scenarioreduction}.
\begin{figure}[t]
% \setlength{\belowcaptionskip}{-10pt}
% \vskip 0.2in
\begin{center}
\centerline{\includegraphics[width=\columnwidth]{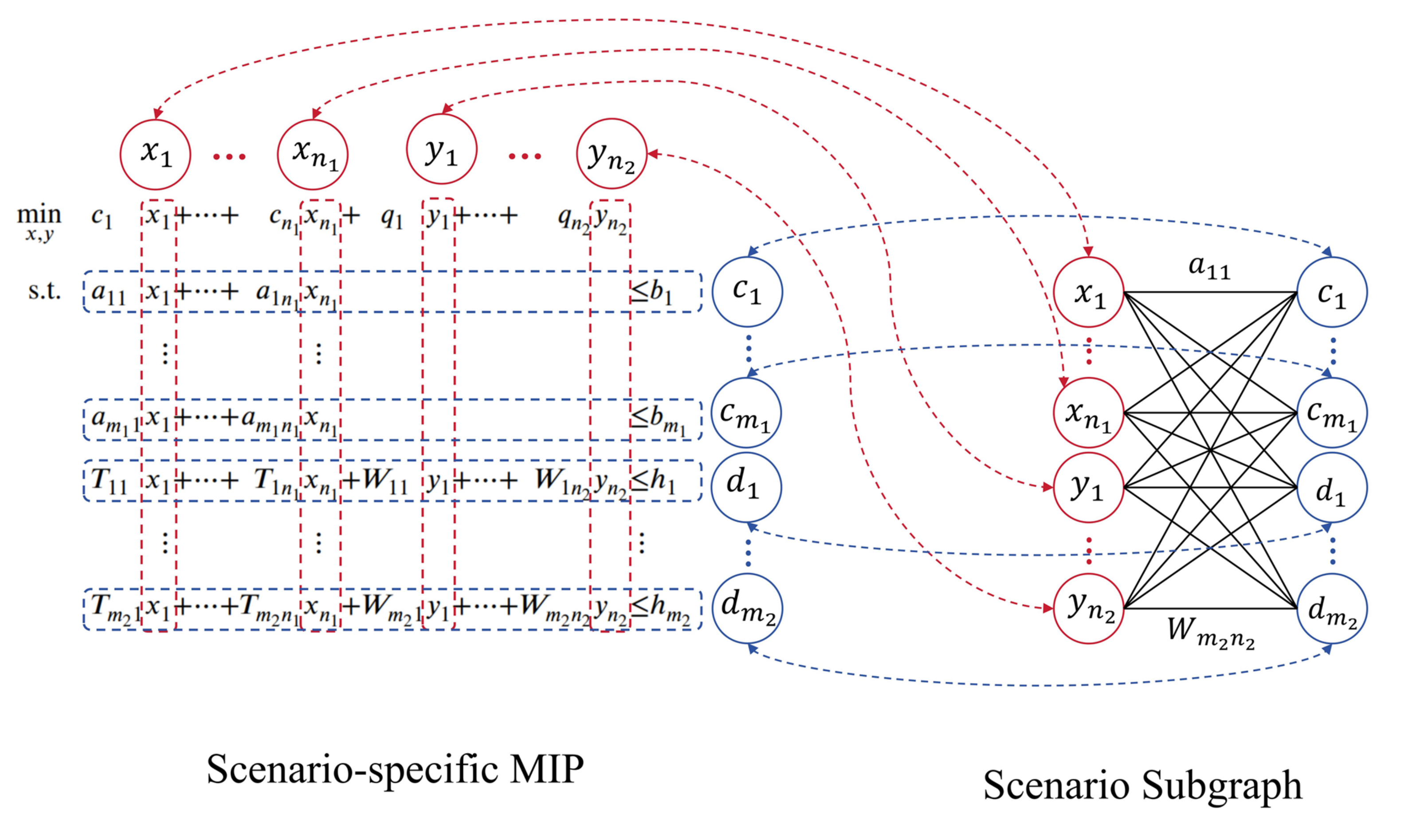}}
\caption{The scenario-specific MIP is represented as a bipartite graph with two sets of nodes. On the left side are the variables $\{x_1, \cdots, x_{n_1}, y_1,\cdots, y_{n_2}\}$, and on the right are the constraints $\{c_1,\cdots,c_{m_1},d_1,\cdots,d_{m_2}\}$.}
\label{fig:scenario_subgraph} 
\end{center}
\vskip -0.3in
\end{figure}
\subsection{Bipartite Graph Representation} \label{subsec:bip_graph}
Each scenario of a 2SP symbolizes a complex MIP with substantial information. To capture its significant attributes, we adopt the weighted bipartite graph representation, as introduced in \cite{gasse2019learn2branch}. It has played a key role in tasks like branch selection, cutting plane selection, and integer programming optimization \cite{gasse2019learn2branch, turner2023adaptive, ye2023gnn}.  As depicted in  \cref{fig:scenario_subgraph}, variables and constraints are distinct nodes. Edges link nodes of different types, with the weight corresponding to the variable's coefficient in the constraint. Important MIP features, such as objective function and variable bounds, are encoded as node features, ensuring a detailed representation of the MIP with minimal loss \cite{Chen2023OnRM, ye2023gnn}. 

\section{Methodology} \label{lab:}
In this section, we outline the proposed HGCN2SP. The framework is summarized in \cref{fig:pipeline}. Consider a set of 2SP instances, denoted as $\{(x^l, P_{l}, \{\xi_{l,j}\}_{j}^{N_l}, v^l)\}_{l}$. Here, $P^l$ and $\{\xi_{l,j}\}_{j}^{N_l}$ represent the deterministic and uncertain parameters of the $l$-th instance, respectively. The optimal value and solution for the $l$-th instance are $v^l$ and $x^l$. 
\par Our goal is to choose scenarios that closely align with the optimal decision while shortening solving time. To achieve this, we construct a hierarchical graph for each instance and then apply a hierarchical graph convolutional network to derive embeddings. An attention-based decoder then uses these embeddings to select representative scenarios. A constant $k \in \mathbb{Z}_{+}$ is set to the number of scenarios to select.

\subsection{Scenario-based Hierarchical Graph}
For each instance $\{(x^l, P_{l}, \{\xi_{l,j}\}_{j}^{N_l}, v^l)\}_{l}$, we represent it as a hierarchical graph. The lower level involves combining the uncertain parameters of a certain scenario with the deterministic information of the first stage, resulting in a scenario subproblem, which is depicted as a bipartite graph (refer to \cref{subsec:bip_graph}). Formally, the combination of $P_l$ and $\{\xi_{l,j}\}_{j}^{N_l}$ can be transformed into $\{G_{l,1}, \dots, G_{l, N_l}\}$, with each $G_{l,j} = (\mathcal{V}_{l,j}, \mathcal{E}_{l,j}, A_{l,j}, X_{l,j})$ being the $j$-th \textit{scenario subgraph}. It includes a feature matrix $X_{l,j} \in \mathbb{R}^{|\mathcal{V}_{l,j}| \times F}$ and an adjacency matrix $A_{l,j} \in \mathbb{R}^{|\mathcal{V}_{l,j}| \times |\mathcal{V}_{l,j}|}$.  

On the higher level, the \textit{instance graph} deems each scenario as a vertex. Edges are weighted based on the similarity of their uncertain parameters.  For $l$-th instance, we denote $A_l$ as the adjacency matrix that reflects interrelationships between scenarios. Hence, this instance is represented by $(G_{l,1}, \dots, G_{l, N_l}, A_l)$ without omitting crucial details.

\subsection{Reinforcement Learning Formulation}
Upon reviewing the order information, we note that extracting a sequence of scenario subsets from a finite scenario space leads to an exponentially large number of possibilities. This makes it impractical to obtain a supervisory signal. Consequently, we take the feedback of the solver as a reward and employ reinforcement learning to efficiently exploit it.

\par  We define scenario reduction as a Markov decision process (MDP) \cite{sutton2018reinforcement} consisting of state space $\mathcal{S}$, action space $\mathcal{A}$, reward function $r: \mathcal{S} \times \mathcal{A} \rightarrow \mathbb{R}$, and transition function $\mathcal{P}$. The details are as follows:

\par \textbf{State space $\mathcal{S}$:} The state captures essential information about a 2SP instance. Specifically, for the $l$-th instance, the state $s_l$ is designed by $(G_{l,1}, \dots,G_{l,N_l}, A_l)$.

\par \textbf{Action space $\mathcal{A}$:}  The action space consists of all permutations of $k$ scenarios from the candidate set. Considering the sequence of scenario selection, this leads to an exponential increase in the number of possible actions.

\par \textbf{Reward function $r$: } 
We assess the chosen scenario sequence by incorporating $k$  scenarios into a MIP and utilizing solver feedback.  With state $s$ and action $a$, expressed as $\{a_l^1,a_l^2,\cdots,a_l^k\}$, an advanced solver is used to solved $\min \{c^{T}x+\sum\limits_{i=1}^{k} p_i 
Q(x,\xi_{a_l^i})\}$  to determine the solving time $t_a$ and optimal solution $\tilde{x}^{*}$. We further get $f(\tilde{x}^{*})$ by substituting $\tilde{x}^{*}$ into \cref{eq:2sp_expected}.  
The reward $r$ comprises two parts: the negative solution time ($-t_a$), and the consistency measure (M) between $\tilde{x}^{*}$ and the optimal solution $x^l$. The final reward is calculated as a weighted sum of the two elements, using $\alpha\in(0,1)$ as the weight parameter.

\begin{equation}
    r(s,a) = -(1-\alpha) t_a +\alpha M_a,
    \label{eq:reward}
\end{equation}

where the matching score $M_a$ is calculated as the negative Manhattan distance, represented by the equation $M_a = -\sum\limits_{i} |\tilde{x}_i^{*}-x_i^l|$. A larger value of $M$ indicates a higher similarity between two decisions. 
\par \textbf{Transition function $\mathcal{P}$: } In our single-step RL model, the transition function relates the initial state and actions to a smaller 2SP containing $k$ scenarios.

\subsection{Policy Network Architecture}
Let $\pi$ denote the scenario selection policy $\pi: \mathcal{S} \rightarrow \mathcal{A}$, with $\pi(\cdot|s)$  indicating the probability distribution over actions given state $s$. 
We employ a sequence-to-sequence architecture to extract problem features and generate sequential outputs. 
%In this context,  we set a constant $k$ for the number of scenarios to select.
Our policy is defined as the parametric model $\pi_{\theta}$ with the  probability of an action $a$ determined as follows:
\begin{equation}
    \pi_{\theta}(a|s) = \prod \limits_{i=1}^{k} p_{\theta}(a^{i}|a^{1},\cdots,a^{i-1}; s).
\end{equation}
The encoder in our policy is a hierarchical GCN that derives embeddings from the hierarchical graph, while the decoder employs an attention-based sequence model.
\par 
\textbf{Hierarchical Graph Convolutional Network} Considering the two-stage nature of 2SP, hierarchical processing is key for capturing fine-grained details within each scenario and exploiting the topological relationships in the scenario space. Therefore, we employ a hierarchical network to process graphs of different levels separately. The network consists of four GCN layers, as proposed by \cite{kipf2016semi}:
\begin{equation}
H^{(k+1)} = \sigma(\tilde{D}^{-\frac{1}{2}}\tilde{A}\tilde{D}^{-\frac{1}{2}}H^{(k)}W^{(k)}) .   
\label{eq:gcn}
\end{equation}
In this equation,  $H^{(k)}$ denotes all node embeddings in $k$-th layer and $H^{(0)}$ is the input node features. $\tilde{A}$ is adjacency matrix with added self-loops, defined as $\tilde{A}=A+I$ , where $I$ is the identity matrix.  $\tilde{D}$ is the diagonal degree matrix of $\tilde{A}$, and $W^{(k)}$ is the learnable weight matrix.  
 The activation function $\sigma(\cdot)$ is chosen to be tanh in our setting. 

\par Though structurally similar, the initial two layers primarily 
aggregate fine-grained information from each scenario subgraph.
Initially, $H^{(0)}$ represents the feature $X_{l,j}$,  which includes properties mainly related to constraints and variables.  Details of these features can be found in \cref{sec:implement_details}.
 \par 
 We define $H_{l,j}\in \mathbb{R}^{ |\mathcal{V}_{l,j}|\times F_1}$   as the high-level node representations obtained by the first two layers. A readout function $\mathcal{R}: \mathbb{R}^{|\mathbb{V}_{l,j}| \times F_1} \rightarrow \mathcal{R}^{F_1}$ is then applied to derive graph-level representations, representing each scenario.
\par  At this point, we have the $l$-th instance graph $(H_l^{0}, A_l)$, where $H_l^{0} \in \mathbb{R}^{N_l\times F_1}$ is the feature matrix of  $l$-th instance. Each entry $H_{l,j}^{(0)}$  represents the $j$-th scenario, which is calculated by $H_{l,j}^{(0)} = \mathcal{R}(H_{l,j})$.
Then, $H_l^{(0)}$ and $A_l$ are fed into the final two GCN layers to obtain the ultimate representation $H_l=\left[h_{l,1},h_{l,2},\cdots,h_{l, N_l} \right]\in \mathbb{R}^{N_l\times F^{'}}$. We also compute the global embedding $\overline{h}_l$ as $\overline{h}_l=\frac{1}{N_l}\sum\limits_{j=1}^{N_l} h_{l,j}$. Both $H_l$ and $\overline{h}_l$ serve as inputs for the decoder. 

\textbf{Attention-based Decoder}  We propose using the Attention Model's decoder,  as introduced by \cite{Kool2018AttentionLT}, for sequentially selecting $k$ scenarios. At each timestep $t\in\{1,2,\cdots,k\}$, the decoder chooses node $\pi_{t}$ based on the global embedding $\overline{h}_l$ and prior outputs $\pi_{t^{'}}$ where $t^{'}<t$. This decoder is unique in that it creates a special context embedding $h_c^{t}$ for attention calculation, thus avoiding extensive $n\times n$ attention computations. The context embedding $h_c^{t}$ comprises the embedding of the global, the previous node $\pi_{t-1}$ and the first node $\pi_1$. In our model, for a more detailed characterization of the selected nodes, we replace the embedding of $\pi_1$ with $h_a^t$, a combination of $h_c^{t-1}$ and $h_a^{t-1}$, and $h_a^{1} = h_{\pi_1}$. The specific definition is as follows:
\begin{equation} 
\begin{aligned}
h_{a}^t = \frac{1}{2} (h_c^{t-1}+h_{a}^{t-1}),\\
    h^{t}_{c} = 
\begin{cases} 
  [\overline{h}, h_{\pi_{t-1}}, h^{t}_{a}] & \text{if } t > 1, \\
  [\overline{h}, v^{f}, v^{f}] & \text{if } t = 1,
\end{cases}
\end{aligned}
\end{equation}
where $[\cdot,\cdot,\cdot]$ is the horizontal concatenation operator, and $v^{l}$ and $v^{f}$ are learnable parameters serving as placeholders.
More details about the model are provided in \cref{sec:decoder}.
\par 
\textbf{Training}  To train the policy network, we adopt Proximal Policy Optimization (PPO) \cite{schulman2017proximal}, notable for its data efficiency and reliable performance. Such a framework utilizes an actor-critic architecture. The actor is the policy network $\pi_{\theta}$. The critic $v_{\phi}$ shares the same encoder with the policy but employs self-attention and a Multi-Layer Perceptron (MLP) as the decoder.  $v_{\phi}$ outputs a scalar value estimating the reward of a state.

\section{Experiments}\label{label:experiment}
\begin{table*}[t]
\caption{Comparison results with baselines on CFLP$\_10\_20\_200$ and NDP$\_2\_2\_10\_200$ with error rates ($\%$) and time in seconds (s). The dataset size for each problem is 100.  We report the mean value of error and time with standard deviation on five seeds. \textbf{Bold means the best result of learning-based methods.} }
\label{tab:cflp_10_20_200}
\vskip 0.1in
\begin{center}
\begin{small}
\setlength\tabcolsep{0.9pt}
\begin{sc}
\begin{tabular}{l|cc|cc|cc|cc|cc|cc}
\toprule
\multirow{3}{*}{Method}& \multicolumn{6}{c|}{CFLP$\_10\_20\_200$} & \multicolumn{6}{c}{NDP$\_2\_2\_10\_200$}\\
 \cline{2-13}
 & \multicolumn{2}{c}{$k$=5} & \multicolumn{2}{|c}{$k$=10} & \multicolumn{2}{|c}{$k$=20} & \multicolumn{2}{|c}{$k$=5} & \multicolumn{2}{|c}{$k$=10} & \multicolumn{2}{|c}{$k$=20}\\
 & Error  & Time& Error & Time& Error & Time & Error & Time& Error & Time& Error & Time\\
\hline
Gurobi    & 0.00 & 10695.66 & 0.00 & 10695.66 & 0.00 & 10695.66 & 0.00& 26.54 & 0.00 & 26.54 & 0.00 & 26.54\\
CSSC & -& - & - & - &- & -& 2.87 & 573.75 & 2.00 & 722.38 & 1.42 & 762.24 \\

NN-P & 21.98 & 521.38 & 21.98 & 521.38 & 21.98 & 521.38 & 37.25 & 3288.99 & 37.25 & 3288.99 & 37.25 & 3288.99\\

NN-E & 32.38 & 3903.09 & 32.38 & 3903.09 & 32.38 & 3903.09 & 66.99 & 7283.30 & 66.99 & 7283.30 & 66.99& 7283.30        \\

CVAE-SIP& 6.60& 2.60 & 3.66 & 12.38 & 1.77 &\textbf{26.12} & 58.62 & 0.57 & 11.06& 1.01 &4.63&1.96\\

CVAE-SIPA& 6.86&  2.40 & 3.49 & \textbf{10.23} & 2.50 &27.55 &\textbf{24.59} &0.54 & \textbf{6.06} &1.01 &\textbf{2.12} &2.37 \\

\multirow{2}{*}{HGCN2SP}  & \textbf{2.47} & 2.45  & \textbf{1.37}& 14.72 & \textbf{1.16} & 41.97 & 33.03 & \textbf{0.17} & 8.10 & \textbf{0.40} &2.25 &\textbf{0.98}\\
&  \textbf{($\pm$ 0.33)} & ($\pm$ 0.37)  & \textbf{($\pm$ 0.03)}& ($\pm$ 1.35) & \textbf{($\pm$ 0.01)} & ($\pm$ 2.72) & ($\pm$ 7.16) & \textbf{($\pm$ 0.01))} &($\pm$1.60) & \textbf{($\pm$ 0.05)} &($\pm$ 0.31) &\textbf{($\pm$ 0.21)}\\ 
\bottomrule
\end{tabular}
\end{sc}
\end{small}
\end{center}
% \vskip 0.1in
\end{table*}

\begin{table*}[t]
\caption{Generalization to larger-scale problems on CFLP$\_20\_40\_200$ and NDP$\_2\_2\_20\_200$ with error rates ($\%$) and time in seconds (s). The dataset size is 50 for CFLP and 100 for NDP.  We report the mean value of error and time with standard deviation on five seeds. \textbf{Bold means the best result of learning-based methods.} }
\label{tab:across_large_instance}
\vskip 0.1in
\begin{center}
\begin{small}
\setlength\tabcolsep{0.9 pt}
\begin{sc}
\begin{tabular}{l|cc|cc|cc|cc|cc|cc}
\toprule
\multirow{3}{*}{Method} & \multicolumn{6}{|c}{CFLP$\_20\_40\_200$} & \multicolumn{6}{|c}{NDP$\_2\_2\_20\_200$}\\
 \cline{2-13}
& \multicolumn{2}{c|}{$k$=5} & \multicolumn{2}{c|}{$k$=10} & \multicolumn{2}{c|}{$k$=20} & \multicolumn{2}{c|}{$k$=5} & \multicolumn{2}{c|}{$k$=10} & \multicolumn{2}{c}{$k$=20}\\
 & Error  & Time& Error & Time& Error & Time & Error & Time& Error & Time& Error & Time\\
\hline
Gurobi   & 0.00 & 10800.69 & 0.00 & 10800.69 & 0.00 & 10800.69 & 0.00&157.90 & 0.00 & 157.90 & 0.00 & 157.90\\
CVAE-SIP& 6.70 & \textbf{15.80} &  2.92 & 145.20 & -0.79 & 938.48 &
72.86 &0.97 & 8.21 & 2.17 & 5.22 & 4.59\\
CVAE-SIPA& 4.54 &  23.26 & 1.50 & 157.07 & -0.88 & 909.40
& \textbf{31.12} & 1.04  & \textbf{7.83} & 2.66 & 2.68 & 5.05\\

\multirow{2}{*}{HGCN2SP} & \textbf{3.97} & 24.20 & \textbf{-0.63} & \textbf{124.71} & \textbf{-2.65} & \textbf{784.90} & 32.21& \textbf{0.41} & 10.62 & \textbf{1.01} & \textbf{2.04} & \textbf{2.22} \\

& \textbf{($\pm$ 0.25)} & ($\pm$ 1.00) & \textbf{($\pm$ 0.47)} & \textbf{($\pm$ 30.12)} & \textbf{($\pm$ 0.43)} & \textbf{($\pm$ 47.86)} & ($\pm$ 5.85) & \textbf{($\pm$ 0.06)} & ($\pm$ 1.30) & \textbf{($\pm$ 0.15)} & \textbf{($\pm$ 0.60)} & \textbf{($\pm$ 0.71)} \\
\bottomrule
\end{tabular}
\end{sc}
\end{small}
\end{center}
% \vskip -0.1in
\end{table*}

All experiments are conducted on a server with an Intel Xeon CPU Gold 5220 @ 2.20GHz, complemented by NVIDIA TITAN RTX GPUs with 24 GB of RAM each. The used solver is Gurobi 10.0.3 \cite{gurobi2021gurobi}. Additionally,  our model relies on Pytorch 2.0.1 \cite{paszke2019pytorch} and PyG 2.4.0 \cite{Fey/Lenssen/2019}. Our code and other resources are available at \texttt{https://github.com/samwu-learn/HGCN2SP/}.
\par \textbf{Datasets} We focus on two classical problems: the Capacitated Facility Location Problem (CFLP) \cite{cornuejols1991comparison, ntaimo2005million}, and the Network Design Problem (NDP) \cite{riis2002capacitated, santoso2005stochastic}. 
\par CFLP, a well-researched topic in literature, involves selecting facilities in the first stage and assigning customers to these facilities in the second stage.   Prior works typically fall into two categories: demand$\_$CFLP \cite{bieniek2015note, patel2022neur2sp}, where customer demands are randomized, and presence$\_$CFLP \cite{wu2022learning, keutchayan2023problem}, focusing on the random presence of customers. Our model is trained in the presence setting. But for combining both, we add randomness of demands to presence.
% Additionally, to showcase our method's adaptability, we test it across all these settings.  
\par NDP deals with transporting various types of commodities from the source to the sink via a directed network.  Decisions about opening network edges need to be made before determining the demand for these commodities. 
\par In the following sections, CFLP instances are denoted as CFLP$\_m\_n\_s$, and NDP instances as NDP$\_o\_d\_n\_s$. Here $m, n, o, d$ and $s$ represent facilities, customers,  sources, sinks, intermediates, and scenarios, respectively.
\par \textbf{Instance generation} We collect 512 instances for CFLP$\_10\_20\_200$, and 2048 for NDP$\_2\_2\_10\_200$. Each instance is solved within 3 hours, which is limited to the solver. See the \cref{sec:instance_generation} for specific information.
%We expand our CFLP data by scaling objective functions and constraints with constants between 0.9 and 1.1, resulting in a total of 8192 CFLP instances for training. 
\par \textbf{Implementation details} We use Adam as the optimizer with an initial learning rate of $2.5\times 10^{-4}$ and a weight decay of $10^{-4}$. For CFLP, we choose $k=5$ scenarios, and for NDP $k=10$ is selected due to its faster computation time. More details are noted in \cref{sec:implement_details}.
\begin{table*}[t]
\caption{Generalization to larger-scenarios problems on CFLP$\_10\_20\_500$ and NDP$\_2\_2\_10\_500$ with error rates ($\%$) and time in seconds (s). The dataset size for each problem is 50.  We report the mean value of error and time with standard deviation on five seeds. \textbf{Bold means the best result of learning-based methods.}}
\label{tab:across_large_scenarios}
\vskip 0.1in
\begin{center}
\begin{small}
\setlength\tabcolsep{0.9 pt}
\begin{sc}
\begin{tabular}{l|cc|cc|cc|cc|cc|cc}
\toprule
\multirow{3}{*}{Method} & \multicolumn{6}{|c|}{CFLP$\_10\_20\_500$} & \multicolumn{6}{|c}{NDP$\_2\_2\_10\_500$}\\
 \cline{2-13}
& \multicolumn{2}{c|}{$k$=5} & \multicolumn{2}{c|}{$k$=10} & \multicolumn{2}{c|}{$k$=20} & \multicolumn{2}{c|}{$k$=5} & \multicolumn{2}{c|}{$k$=10} & \multicolumn{2}{c}{$k$=20}\\
 & Error  & Time& Error & Time& Error & Time & Error & Time& Error & Time& Error & Time\\
\hline
Gurobi    & 0.00 & 21601.34 & 0.00 & 21601.34 & 0.00 & 21601.34 & 0.00&537.71 & 0.00 & 537.71 & 0.00 & 537.71\\

CVAE-SIP& 10.18 & 3.41& 4.05 & 14.02 & 2.20 & 72.60
& 53.70 & 0.76 &10.02 & 1.14 & 4.46 & 2.01 \\

CVAE-SIPA& 11.53 & 3.32  & 5.37 & \textbf{11.71} & 2.00 & \textbf{22.33}
& 30.54 & 0.76 & \textbf{8.15} & 1.29 & 3.03  & 2.08 \\

\multirow{2}{*}{HGCN2SP}   & \textbf{3.43} & \textbf{3.16} & \textbf{1.70} & 15.00 & \textbf{1.52}& 26.76 & 
\textbf{29.84} & \textbf{0.21} & 8.17 & \textbf{0.40} & \textbf{2.85} & \textbf{1.00} \\

& \textbf{($\pm$ 0.05)} & \textbf{($\pm$ 0.06)} & \textbf{($\pm$ 0.02)} & ($\pm$ 1.73)& \textbf{($\pm$ 0.03)}& ($\pm$ 1.46) & 
\textbf{($\pm$ 5.32)} & \textbf{($\pm$ 0.07)} & ($\pm$ 1.23) & \textbf{($\pm$ 0.09)} & \textbf{($\pm$ 0.76)} & \textbf{($\pm$ 0.16)} \\
\bottomrule
\end{tabular}
\end{sc}
\end{small}
\end{center}
\vskip 0.1in
\end{table*}

\begin{table*}[t]
\caption{Ablation study of hierarchical graph model on CFLP$\_10\_20\_200$ and NDP$\_2\_2\_10\_200$ with error rates ($\%$) and time in seconds (s). The dataset size for each problem is 100. \textbf{Bold means the best result of learning-based methods.}}
\label{tab:hgcn_ab}
\vskip 0.1in
\begin{center}
\begin{small}
\setlength\tabcolsep{2 pt}
\begin{sc}
\begin{tabular}{l|cc|cc|cc|cc|cc|cc}
\toprule
\multirow{3}{*}{Method} & \multicolumn{6}{|c}{CFLP$\_10\_20\_200$} & \multicolumn{6}{|c}{NDP$\_2\_2\_10\_200$}\\
 \cline{2-13}
& \multicolumn{2}{|c}{$k$=5} & \multicolumn{2}{|c}{$k$=10} & \multicolumn{2}{|c}{$k$=20} & \multicolumn{2}{|c}{$k$=5} & \multicolumn{2}{|c}{$k$=10} & \multicolumn{2}{|c}{$k$=20}\\
 & Error  & Time& Error & Time& Error & Time & Error & Time& Error & Time& Error & Time\\
\hline
%EF    & 0.00 & 21601.34 & 0.00 & 21601.34 & 0.00 & 21601.34 & 0.00&537.71 & 0.00 & 537.71 & 0.00 & 537.71\\

CVAE-SIP$_{small}$& 7.67& 2.84 & 4.90 & \textbf{10.55} & 3.74 & 27.25 & 43.19 & 0.37 & 16.61& 0.86 &5.07&1.79\\

CVAE-H$_{small}$&  \textbf{6.11}& \textbf{2.66} & \textbf{2.93}& 10.68 & \textbf{1.81} & \textbf{24.98}
& \textbf{40.46} & \textbf{0.35} &\textbf{14.37} &\textbf{0.83} & \textbf{3.98} & \textbf{1.77} \\

% Our  & \textbf{1.89} & \textbf{2.33} & \textbf{1.35}& 14.89 & \textbf{1.14} & \textbf{42.45} & \textbf{23.99} & \textbf{0.15} & \textbf{6.68} & \textbf{0.33} &\textbf{2.86} &\textbf{0.83}\\
\bottomrule
\end{tabular}
\end{sc}
\end{small}
\end{center}
\vskip -0.1in
\end{table*}

\textbf{Baselines} In experiments, we compare our method with the following baselines and limit the number of threads to 16: \par 
1) \textbf{Gurobi:} Utilizes Gurobi for optimal EF values. The time limits are set as 3 hours, except for CFLP$\_10\_20\_500$, where the time bound is extended to 6 hours. \par 
2) \textbf{CSSC \cite{keutchayan2023problem} :}  A scenario reduction approach with a theoretical guarantee. It calculates the optimal values of every individual scenario and constructs a MIP in the value space for choosing representative scenarios. \par
3) \textbf{NN-P and NN-E \cite{patel2022neur2sp} :}  These methods use a Rectified Linear Unit (ReLU) neural network to approximate the objective of a set of scenarios and then convert the network into a MIP for optimization. \par
4) \textbf{CVAE-SIP \cite{wu2022learning} :} Employs a conditional variational autoencoder to learn scenario representations and clusters them to identify representative scenarios. \textbf{CVAE-SIPA} further enhances representations by also predicting the optimal value of individual scenarios.

\subsection{Comparison Analysis}
For each problem, we test on 100 instances, using $k=5$, 10, and 20 to evaluate scenario reduction algorithms. We remind that methods like NN-P, NN-E, and EF are not affected by the choice of $k$. Besides, due to limited generalization, NN-P and NN-E require instance-specific training for each instance.  Therefore, we only test them on 10 instances and record their actual time (including training and evaluation).

\cref{tab:cflp_10_20_200} demonstrates that HGCN2SP significantly outperforms other learning methods on the CFLP dataset.  Particularly noteworthy is our method's performance with $k=5$. It takes only 2.45 seconds to achieve a result, with a gap of a mere $2.47\%$, surpassing CVAE-SIPA even when it selects 20 scenarios. Furthermore, as the number of selected scenarios increases, the performance of HGCN2SP improves,  but so does the time required for a solution.  Still,  it remains significantly less than the 3 hours needed by the Gurobi,  highlighting the efficiency of our method. On the NDP dataset, our method underperforms compared to CVAE-SIPA, yet we get the first-stage decision in a much shorter time,  taking less than half the time of the CVAE methods. Especially for $k=20$, it takes only 0.98 seconds to achieve comparable results to CVAE, which takes 2.37 seconds.

The above results indicate that HGCN2SP has minor shortcomings on the NDP dataset. The primal reason is the model's reward dependence on aligning the first-stage decision with the actual solution. For the CFLP, with a manageable number of decision variables (equal to the number of facilities, such as 10 for CFLP$\_10\_20$), enabling effective model learning. In contrast, the NDP$\_2\_2\_10$, with only 14 vertices but using edges as decision variables, results in a total of 178 variables. Consequently, minor changes in variables do not significantly affect the reward. However, the objective value of NDP is highly sensitive to these decisions, thus imposing limitations.

NN-P and NN-E combine scenario parameters and first-stage decisions for predictions, later converting the neural network into a MIP problem.  The two methods have limited performance for two main reasons. First, they use shallow ReLU networks for prediction, which become inaccurate with large input dimensions (230 for CFLP$\_10\_20$ and 182 for NDP$\_2\_2\_10$). Second, the size of the MIP problem, created by transforming the neural network, grows in size linearly with the number of nodes and layers of the network, leading to a slower solving time.

\subsection{Generalization across larger-scale problems}
In practical applications, a model's ability to handle larger-scale problems is essential. To verify the capabilities of HGCN2SP, we apply the trained models directly to 50 CFLP and 100 NDP larger-scale instances. The CFLP instances have twice the number of facilities and customers, while the NDP instances double the intermediate nodes. According to \cref{tab:across_large_instance},  our model remains effective on these larger instances. Notably, despite spending three hours, the Gurobi still can not solve the enlarged CFLP instance. With 10 selected scenarios, our method achieves $0.63\%$ better results than Gurobi in 124.71 seconds. Compared to other learning methods, HGCN2SP obtains better solutions for all three values of $k$ and leads in time for $k=10$ and $k=20$. In the NDP, HGCN2SP yields results superior to CVAE at $k=20$ but falls short when $k$ is 5 and 10. Remarkably, our results are achieved in less than half the time required by CVAE. Besides, the solving time for the Gurobi increases from 26.54 seconds to 157.90 seconds, complicating its practical application. In contrast, our method makes decisions with minimal difference in just 2.22 seconds. These performances demonstrate our method's strong generalization on larger-scale problems and its advantage in solving time.

\subsection{Generalization across larger scenarios}

Scenarios depict the uncertainty of the future, which may shift with new information or the failure of previous predictions. \cite{issac2017shortest}. Since such situations are common in practice, the adaptability to varying scenario sizes is crucial. Therefore, we conduct tests on 50 instances, each with 500 scenarios. As \cref{tab:across_large_scenarios} indicates, HGCN2SP exhibits strong generalization in large-scale scenarios. In the CFLP, it exceeds baselines by a large margin. Particularly at $k=5$, its performance is 3.43$\%$,  which is $6.75\%$ better than the comparison methods. In the NDP, our method beats all baselines at $k=5$ and 20. When $k$ is 10, our method trails by a mere $0.02\%$. Interestingly, such superior performance has not been observed in NDP$\_2\_2\_10\_200$ and NDP$\_2\_2\_20\_200$, showcasing outstanding generalization ability of our method on larger scenario cardinalities. Moreover, HGCN2SP achieves superior results in less than half the time needed by CVAE. 
\par  As evidenced by the time Gurobi takes, dealing with up to 500 scenarios is highly challenging. For the CFLP, the solver struggles to fully solve the problem even within 6 hours, whereas our approach is only $1.52\%$ off the optimal solution and takes just 26.76 seconds. A similar pattern occurs in NDP, emphasizing the efficiency and practicality of HGCN2SP for real-world applications.

\begin{figure}[t]
\vskip 0.1in
  \begin{center}
      
  % Subfigure a
  \begin{minipage}{0.45\linewidth}
    \centering
    \includegraphics[width=\linewidth]{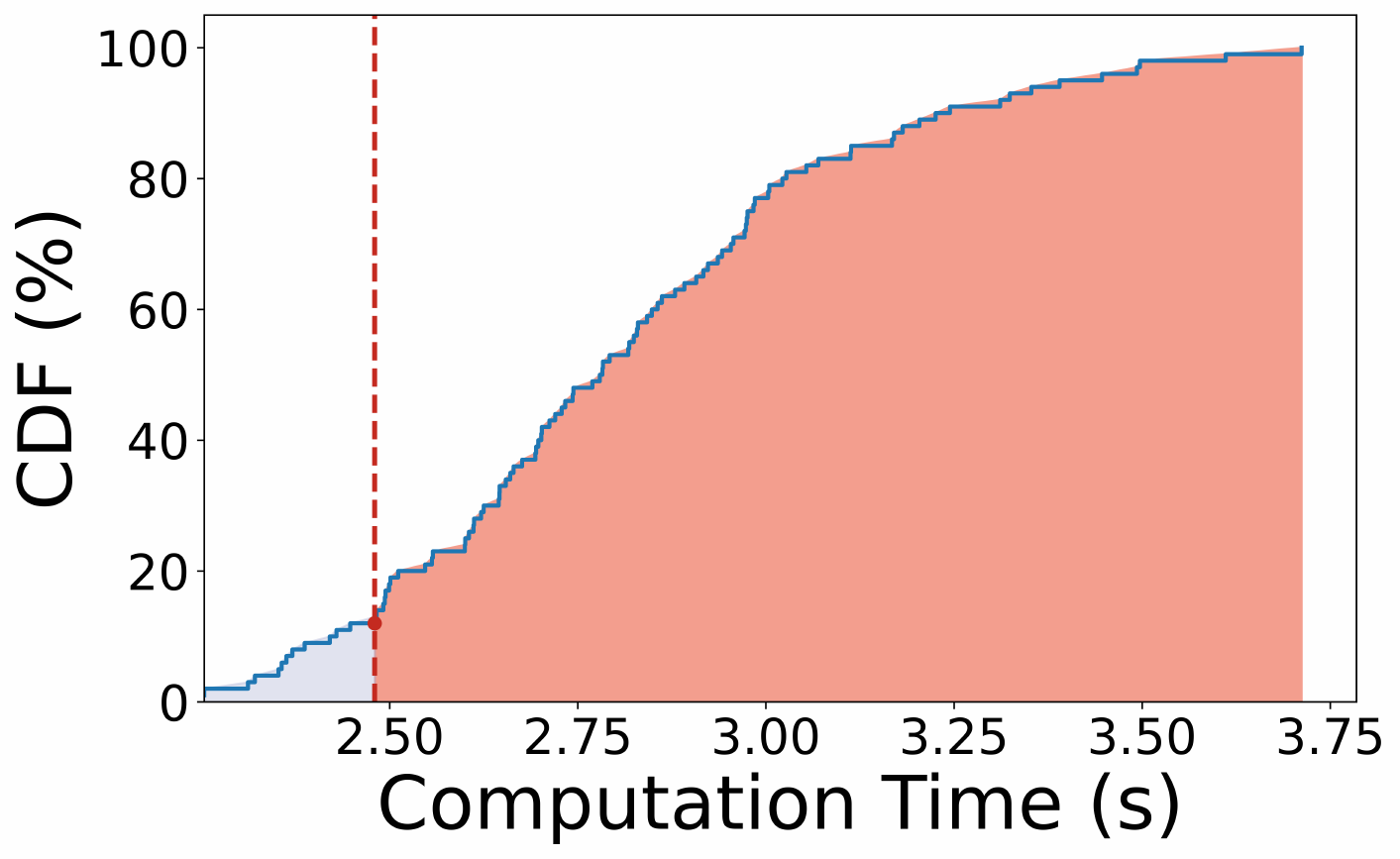}
    % Manually add the subcaption (a)
    \textbf{(a)} $k=5$
    \label{fig:subfig_a} % This label is not referenced if subcaption is not used
  \end{minipage}
  \hfill
  % Subfigure b
  \begin{minipage}{0.45\linewidth}
    \centering
    \includegraphics[width=\linewidth]{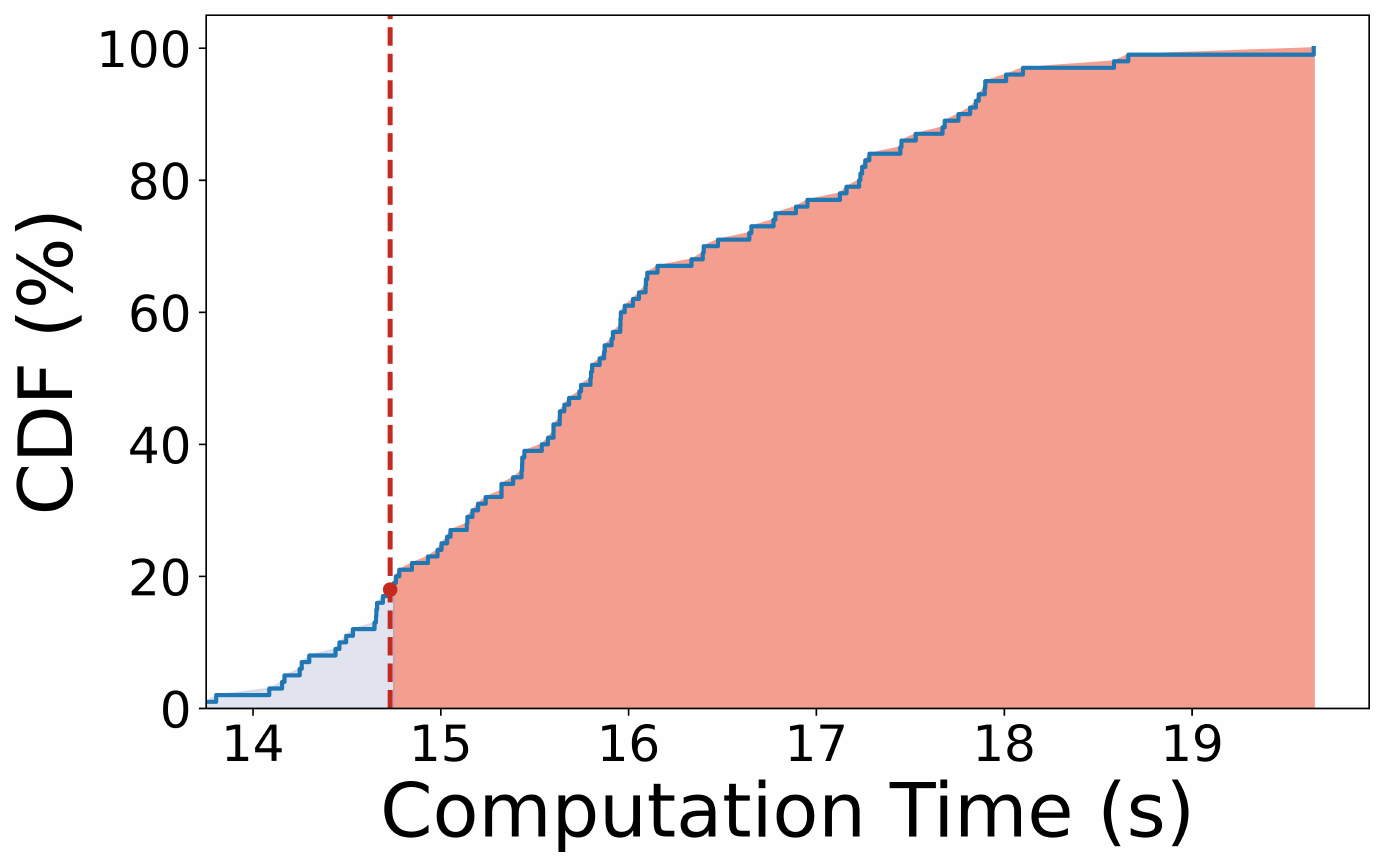}
    % Manually add the subcaption (b)
    \textbf{(b)} $k=10$
    \label{fig:subfig_b} % This label is not referenced if subcaption is not used
  \end{minipage}

  \centering
  \includegraphics[width=1\linewidth]{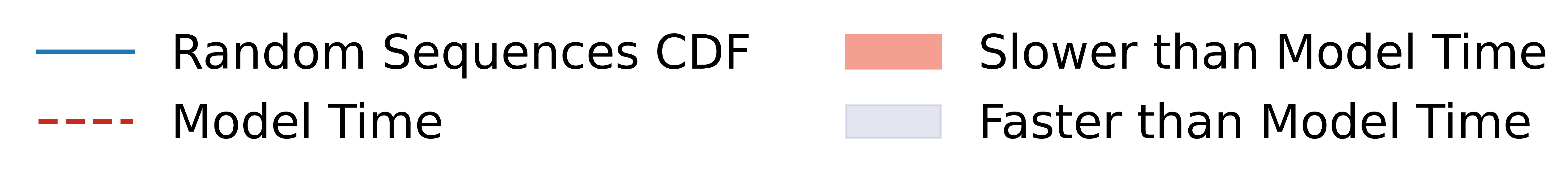}

  \caption{
The model's output is randomized using 100 seeds, with solving times plotted on a cumulative distribution function (CDF). The model's actual solving time, marked on the CDF, outperforms most randomized sequences.}
  \label{fig:cdf_of_time_eff}
  
  \end{center}
  \vskip -0.1in
\end{figure}

\subsection{Time Efficiency}
In our reinforcement learning training, we consider the solving time and the consistency of decision variables for the reward. For assessing the time efficiency of HGCN2SP in scenario selection, we conduct experiments on CFLP$\_10\_20\_200$. We set 100 random seeds to shuffle the output sequence of the model for each instance and then measure the solving time with the solver. To counteract randomness and hardware variability, we average the results over two repetitions for each seed. As \cref{fig:cdf_of_time_eff} illustrates, the sequence ordered by our model greatly outperforms random ones. When 5 scenarios are selected,  only $12\%$ of the random sequences achieve a faster solving time. With 10 scenarios, our model is slower than only $18\%$ of the random sequences despite a significant increase in possible permutations. The results confirm that our model can arrange the selected scenarios better to accelerate the solver's process.

\subsection{Hierarchical Graph Model}
In our paper, we propose a novel hierarchical graph model specifically tailored to the unique characteristics of 2SP problems. This is complemented by an innovative hierarchical graph convolutional network that effectively processes scenarios. The low-level network extracts features within each scenario subgraph, while the high-level network leverages the topological relationships across scenarios to enhance such features. This approach is distinct from CVAE-SIP and CVAE-SIPA, which rely on complete graphs appropriate only for graph-based 2SP problems and fail to consider inter-scenario correlations. 

To showcase the effectiveness of our hierarchical graphs and neural networks, we incorporate this model into CVAE-SIP,  forming a new model CVAE-H. We train it with just 2000 instances on CFLP$\_10\_20\_200$ and NDP$\_2\_2\_10\_200$. To distinguish these from models trained on 12,800 instances, we labeled them with the subscript 'small'. As detailed in \cref{tab:hgcn_ab}, CVAE-H$_{small}$ achieves a $6.11\%$ performance on CFLP with $k=5$, surpassing both CVAE-SIP$_{small}$ and the versions of CVAE-SIP and CVAE-SIPA trained on 12800 instances. Similar outcomes are observed at $k=10$ and $k=20$. In NDP, CVAE-H$_{small}$ also significantly outperform CVAE-SIP$_{small}$, especially at $k=20$, where it also exceeds CVAE-SIP. These results validate our hierarchical graph model's superiority in extracting and leveraging scenario representations for 2SP. Moreover, they reveal the model's ability to considerably boost the performance of other methods when integrated, highlighting its adaptability and potential for performance enhancement.

\section{Conclusions}

In this work, we introduce a novel model, HGCN2SP, for decision-making under uncertainty. Leveraging the hierarchical nature of the two-stage stochastic program, we propose a hierarchical graph model. The low level consists of a bipartite graph representing individual scenarios, while the high level forms an instance graph depicting the overall structure. Furthermore, a hierarchical graph neural network is applied to extract representations of each scenario, which are then fed into an attention-based decoder to select representative ones. During training, we consider not only the performance of selected scenarios but also the impact of ordering on the solving time. Our experiments on two classic 2SPs demonstrate that HGCN2SP can make excellent decisions in less time. Additionally, our method shows strong generalization capabilities in instances with larger scales and larger cardinalities of scenarios.

\section*{Limitations}
Our work has a significant limitation: the collection of training data is time-consuming. For the CFLP problem, accurately solving one instance takes almost 3 hours, and our training process necessitates 512 instances along with their optimal solutions, demanding substantial computational resources. We consider the reduction of training costs a crucial issue that requires immediate attention and a promising direction for future research.
\section*{Impact Statement}
This work proposes a method that uses machine learning to solve two-stage stochastic programming problems. Two-stage stochastic programming is a widely used framework for decision-making under uncertainty, frequently applied in business, science, and technology. Consequently, we expect this work to have a positive impact. However, if used for harmful purposes, it may result in negative social impact.

\section*{Acknowledgements}

This work was supported in part by the National Key R$\&$D Program of China (2022ZD0116402), NSFC 62273347, Jiangsu Key Research and Development Plan (BE2023016).

\nocite{langley00}

\bibliography{example_paper}
\bibliographystyle{icml2024}

%%%%%%%%%%%%%%%%%%%%%%%%%%%%%%%%%%%%%%%%%%%%%%%%%%%%%%%%%%%%%%%%%%%%%%%%%%%%%%%
%%%%%%%%%%%%%%%%%%%%%%%%%%%%%%%%%%%%%%%%%%%%%%%%%%%%%%%%%%%%%%%%%%%%%%%%%%%%%%%
% APPENDIX
%%%%%%%%%%%%%%%%%%%%%%%%%%%%%%%%%%%%%%%%%%%%%%%%%%%%%%%%%%%%%%%%%%%%%%%%%%%%%%%
%%%%%%%%%%%%%%%%%%%%%%%%%%%%%%%%%%%%%%%%%%%%%%%%%%%%%%%%%%%%%%%%%%%%%%%%%%%%%%%
\newpage
\appendix
\onecolumn

\section{The datasets used in evaluation}\label{sec:dataset}
\subsection{ Capacitated Facility Location Problem}
The Capacitated Facility Location Problem (CFLP) is a classic 2SP problem involving a set of facilities $F$ and customers $C$.  Decisions on which facilities to open are made without full knowledge of customer information. Each facility is equipped with resources to satisfy incoming customers.  Generally,  the uncertainty of CFLP involves either the presence or the demand of customers.  In this paper, we explore a more complex variant that combines both types of uncertainty.

The problem is described as follows: In the first stage, the decision-maker needs to decide which facilities to open and is aware of the opening cost $o_f$ and the maximum resources $q_f$ of each facility $f$, as well as the cost for the customer $c$ to access facility $f$. Additionally, we restrict the number of opened facilities to not exceeding $v$.

For each scenario $s$, there are binary random vectors $h^{s} \in \{0,1\}^{|C|}$ and real-valued vectors $q^{s}\in \mathbb{R}^{|C|\times |F|}$. If $h_c^{s} = 1$, the customer $c$ is present; otherwise, they are not considered. The value $q_{cf}^s$ represents the demand of customer $c$ at facility $f$.

In summary, the definition of CFLP with $N$ scenarios is defined as follows:
\begin{align}
\min \quad & \sum_{f \in F} o_{f}x_{f} + \frac{1}{N} \sum_{s=1}^{N} \left( \sum_{c \in C} \sum_{f \in F} s_{cf}y_{cf}^{s} + \sum_{f \in F} b_{f}z_{f}^{s} \right) \nonumber \\
\text{s.t.} \quad & \sum_{f \in F} x_{f} \leq v, \nonumber \\
& \sum_{c \in C} q_{cf}^{s}y_{cf}^{s} \leq q_{f}x_f + z_{f}^{s}, \enspace\quad\quad\quad \forall (f,s) \in F \times \{1, \cdots, N\} \nonumber \\
& z_{f}^{s} \leq Mx_{f}, \,\qquad\qquad\qquad\qquad\quad \forall  (f,s) \in F \times \{1, \cdots, N\} \\
& \sum_{f \in F} y_{cf}^{s} = h_{c}^{s}, \qquad\quad\quad\qquad\qquad \forall  (c,s) \in C \times \{1, \cdots, N\} \nonumber \\
& x_f,y_{cf}^{s}\in \{0, 1\}; z_f^{s} \in [0,\infty), \nonumber
\end{align}
where $x_f$, $y_{cf}^{s}$, and $z_{f}^{s}$ are decision variables. When $x_f=1$, it indicates that the $f$-th  facility is open. If customer c is assigned to f, then $y_{cf}^{s}=1$. In cases where facility $f$ cannot fulfill the demand of the arrived customers,  additional resources $z_{f}^{s}$ are required, with a penalty of $b_f$ per unit.

\subsection{Network Design Problem}
The Network Design Problem (NDP) is a graph-based two-stage stochastic programming problem. It involves designing a network for transporting multiple commodities before knowing their demand.  The initial decision involves determining which edges are linked in the network. Once the demand for various commodities is determined, decisions on commodity flow are made based on the network structure.

The problem is structured as follows:  Given a complete directed graph $G=(\mathcal{V},\mathcal{E})$, containing a set of source nodes $\mathcal{S}$ and sink nodes $\mathcal{T}$,  along with the commodity set $C$. Each edge $e_{ij} \in \mathcal{E}$ includes an opening cost $o_{ij}$ and a capacity $q_{ij}$. For every commodity $c\in C$, the cost per unit transported on this edge is $t_{ijc}$.  The goal is to minimize the total cost of opening edges and transporting all commodities in the network, while also satisfying the demand for goods at each point.  If the demand at a specific point is not met, a penalty $b_p$ will be applied to the source point $p\in \mathcal{S}$. 

In each scenario, the uncertainty is the demands $d^{s}\in \mathbb{R}^{|\mathcal{V}|\times |C|}$, where $d_{vc}^{s}$ is the demand of commodity $c$ at the vertex $v$.

\begin{table}[htbp]
\centering
\setlength\tabcolsep{8pt}

\caption{Features of Scenario subgraph}
\label{tab:feature_of_problem}
\begin{tabular}{|c|c|c|c|c|}
\hline
\multirow{2}{*}{Problem} & \multirow{2}{*}{Feats} & \multirow{2}{*}{Types} & \multirow{2}{*}{Parameters} & \multirow{2}{*}{Dims} \\ 
& & & &\\
  \hline
\multirow{5}{*}{CFLP}&   \multirow{3}{*}{Variable}  & $x_{f}$   &  opening cost $o_f$, capacity $q_f$ & 2           \\  \cline{3-5}
& & $y_{cf}^{s}$ &  demand $q_{cf}^{s}$, transport cost $s_{cf}$    & 2 \\     \cline{3-5}
& & $z_f^s$ &  penalty of per recourse $b_f$ & 1 \\       \cline{2-5}

 & Constraint & parallelism & the parallelism between the objective and the constraint & 1   \\ 
 \cline{2-5}
& Edge& coeff& coefficient of variables in constraints & 1 \\ \hline
 
\multirow{3}{*}{NDP}&   \multirow{2}{*}{Variable}  &   $x_{ij}$                          &  opening cost $o_{ij}$, capacity $q_{ij}$ & 2                 \\  \cline{3-5}
& & $y_{ijc}^{s}$ & transportation cost $t_{ijc}$ & 1 \\ \cline{2-5}

 & Constraint & parallelism &the parallelism between the objective and the constraint  & 1    \\
  \cline{2-5}
& Edge& coeff& coefficient of variables in constraints & 1 \\ \hline
\end{tabular}
\end{table}

The definition of NDP with $N$ scenarios is defined as follows:
\begin{align}
\min \quad & \sum_{e_{ij} \in \mathcal{E}} o_{ij}x_{ij} + \frac{1}{N} \sum_{s=1}^{N} \left( \sum_{c \in C} \sum_{e_{ij} \in \mathcal{E}} t_{ijc}y_{ijc}^{s} + \sum_{p \in \mathcal{S}} b_{p}z_{pc}^{s} \right) \nonumber \\
\text{s.t.} \quad & \sum_{v_i=v, e_{ij}\in \mathcal{E}} y_{ijc}^{s} - \sum_{v_j=v, e_{ij}\in \mathcal{E}} y_{ijc}^{s} = d_{vc}^{s}, 
\enspace \qquad \forall (v,c,i) \in \mathcal{V}\times C \times \{1,\cdots,N\}\nonumber \\
& \sum_{v_i=t, e_{ij}\in \mathcal{E}} y_{ijc}^{s}= 0, 
\qquad \qquad \qquad \qquad \qquad \forall (p,t,c,s)\in \mathcal{S}\times \mathcal{T}\times C \times \{1,\cdots,N\}\\ 
& \sum_{v_j=s, e_{ij}\in \mathcal{E}} y_{ijc}^{s}= 0, 
\qquad \qquad \qquad \qquad \qquad \forall (p,t,c,s)\in \mathcal{S}\times \mathcal{T}\times C \times \{1,\cdots,N\}\nonumber \\ 
& d_{pc}^{s} \leq \sum_{v_i=p, e_{ij}\in \mathcal{E}} y_{ijc}^{s} + M z_{pc}^{s}, 
\qquad \qquad  \quad \quad \forall (p,c,s)\in \mathcal{S}\times C\times \{1,2,\cdots,N\} \nonumber \\
& \sum_{c} y_{ijc}^{s}\leq q_{ij}x_{ij}, 
\qquad \qquad \qquad \qquad \qquad \quad \forall (v_i,v_j,s) \in \mathcal{V}\times \mathcal{V} \times \{1,\cdots,N\}\nonumber\\
& x_{ij}\in \{0,1\};y_{ijc}^{s} \in [0,\infty);z_{pc}^{s}\in \{0,1\}, \nonumber
\end{align}
where $x_{ij}$, $y_{ijc}^{s}$, and $z_{pc}^{s}$ are decision variables. When $x_{ij}=1$, the edge $e_{ij}\in \mathcal{E}$ is open. The variable $y_{ijc}^{s}$ 
  represents volume of commodity $c$ transported on edge  $e_{ij}$. If the commodities shipped by source $p$ are not fully transported due to capacity constraints, a penalty $b_p$ is imposed on source $p$ for commodities not fully transported due to capacity constraints.

\section{Instance Generation} \label{sec:instance_generation}
\textbf{CFLP}: We set the number of facilities and customers to 10 and 20, respectively. Their locations are represented by two-dimensional coordinates, randomly selected from the range [0,1]. Transmission distance is determined using the Euclidean distance. To vary transmission costs, we multiply the distance by a random number from [5, 105]. Facility opening costs and capacities are uniformly sampled from [600, 1500] and [100, 150], respectively. We also limit the maximum number of facilities to 8. If a facility can't meet its customers' total demand, it faces a penalty of 1000 for each lacking resource unit. Customer presence is decided using the Bernoulli distribution, with each customer's probability of presence sampled from [0.8, 0.9]. Customer demands at facilities are uniformly sampled from [20, 80]. Because the EF method for CFLP is time-consuming, we solved the optimal solution of 512 instances. We then expand the dataset to 8192 by scaling the objective function and constraints of these solutions by a factor between [0.9, 1.1]. 

\textbf{NDP}: Following the generation details in \cite{wu2022learning}, the NDP instance includes 14 points: 2 sources and 2 sinks. Commodities shipped from sources are transported through the network to the sinks. The other 10 points are used only for transportation and do not have demand. The number of commodities transported is set to two. In each scenario, commodity quantities are uniformly sampled from [5,15]. Each edge's opening cost, transportation cost, and capacity are randomly selected from [3,11], [5,11], and [10,41], respectively. If a source point's capacity limits full transportation to the sink point, a penalty of 1,000 is applied to that source. We generate 200 scenarios for all instances.

 To evaluate larger-scale problems, we doubled the number of facilities and customers in CFLP and the number of intermediate nodes in NDP. In tests with more scenarios, the number of scenarios is increased to 500.

\begin{table}[ht]
\caption{PPO Hyperparameters used in our experiments}
\label{tab:details_of_ppo}
% \vskip 0.15in
\begin{center}
\setlength\tabcolsep{8pt}

\begin{tabular}{l|l|l|c|c|c}
\toprule
 \multicolumn{3}{l|}{Hyperparameter} & \multicolumn{3}{l}{Value}
 %6}{c}{CFLP$\_10\_20\_200$} & \multicolumn{6}{c}{NDP$\_2\_2\_10\_200$}\\ 
\\
\hline
%EF    & 0.00 & 21601.34 & 0.00 & 21601.34 & 0.00 & 21601.34 & 0.00&537.71 & 0.00 & 537.71 & 0.00 & 537.71\\
 \multicolumn{3}{l|}{Optimizer} & \multicolumn{3}{c}{Adam} \\
 \multicolumn{3}{l|}{Learning rate (actor)} & \multicolumn{3}{c}{$2.5\times 10^{-4}$}\\
 \multicolumn{3}{l|}{Learning rate (critic)} & \multicolumn{3}{c}{$2.5\times 10^{-4}$} \\
 \multicolumn{3}{l|}{Number of environment} & \multicolumn{3}{c}{2048} \\
 \multicolumn{3}{l|}{Number of epochs} & \multicolumn{3}{c}{10 (50 for NDP)}\\
 \multicolumn{3}{l|}{Minibatch size}  &  \multicolumn{3}{c}{16} \\ 
 \multicolumn{3}{l|}{Update epochs} & \multicolumn{3}{c}{10} \\
 \multicolumn{3}{l|}{Clip coefficient} & \multicolumn{3}{c}{0.2} \\
 \multicolumn{3}{l|}{GAE parameter} & \multicolumn{3}{c}{0.95} \\
 \multicolumn{3}{l|}{Vf coefficient} &  \multicolumn{3}{c}{0.5}\\

% Our  & \textbf{1.89} & \textbf{2.33} & \textbf{1.35}& 14.89 & \textbf{1.14} & \textbf{42.45} & \textbf{23.99} & \textbf{0.15} & \textbf{6.68} & \textbf{0.33} &\textbf{2.86} &\textbf{0.83}\\
\bottomrule
\end{tabular}
\end{center}
\vskip -0.1in
\end{table}

\section{Implementation Details and Hyperparameters} \label{sec:implement_details}
\textbf{Features for scenario subgraph}
To encode a single scenario, we construct a bipartite graph representing the MIP of that scenario. The vertices represent decision variables and constraints, with the edge weight between them indicating the variable's coefficient in the constraint. Both variables and constraints carry unique information about the problem and the specific scenario, which are detailed in the \cref{tab:feature_of_problem}.

\textbf{Detail of Instance graph} Once we obtain the deep representation of each scenario, we connect them to form an instance graph. It pairs different scenarios, assigning weights based on the cosine similarity of their scenario uncertainty parameter vectors. Specifically, considering scenarios $\xi_i$ and $\xi_j$, with their corresponding sub-problems $\min\{c^T x + q_{\xi_i}^T y | Ax \leq b, W_{\xi_i} y \leq h_{\xi_i} - T_{\xi_i} x, x \in \mathbb{R}^{p_1} \times \mathbb{Z}^{n_1-p_1}, y \in \mathbb{R}^{p_2} \times \mathbb{Z}^{n_2-p_2}\}$ and $\min\{c^T x + q_{\xi_j}^T y | Ax \leq b, W_{\xi_j} y \leq h_{\xi_j} - T_{\xi_j} x, x \in \mathbb{R}^{p_1} \times \mathbb{Z}^{n_1-p_1}, y \in \mathbb{R}^{p_2} \times \mathbb{Z}^{n_2-p_2}\}$, respectively. For each subproblem, we merge all random variables into a one-dimensional vector $(q_{\xi}, \text{flatten}(W_{\xi}), h_{\xi}, \text{flatten}(T_{\xi}))$, calculate their cosine similarity and use this as the edge weight between scenarios $\xi_i$ and $\xi_j$ in the instance graph.
We then use this graph for feature enhancement within the scenario space.

\textbf{Hyperparameters in PPO}:
This paper employs Proximal Policy Optimization (PPO) as the training framework, with its hyperparameters detailed in the \cref{tab:details_of_ppo}.  Besides,  the $\alpha$ of reward (see \cref{eq:reward}) are 0.001 for CFLP and 0.01 for NDP, respectively.
This work uses the PPO implementation from the CleanRL \cite{cleanrl}.
\begin{figure}[ht]
% \setlength{\belowcaptionskip}{-10pt}
% \vskip 0.2in
\begin{center}
\centerline{\includegraphics[width=0.9\textwidth, trim=0cm 0cm 0cm 0cm, clip]{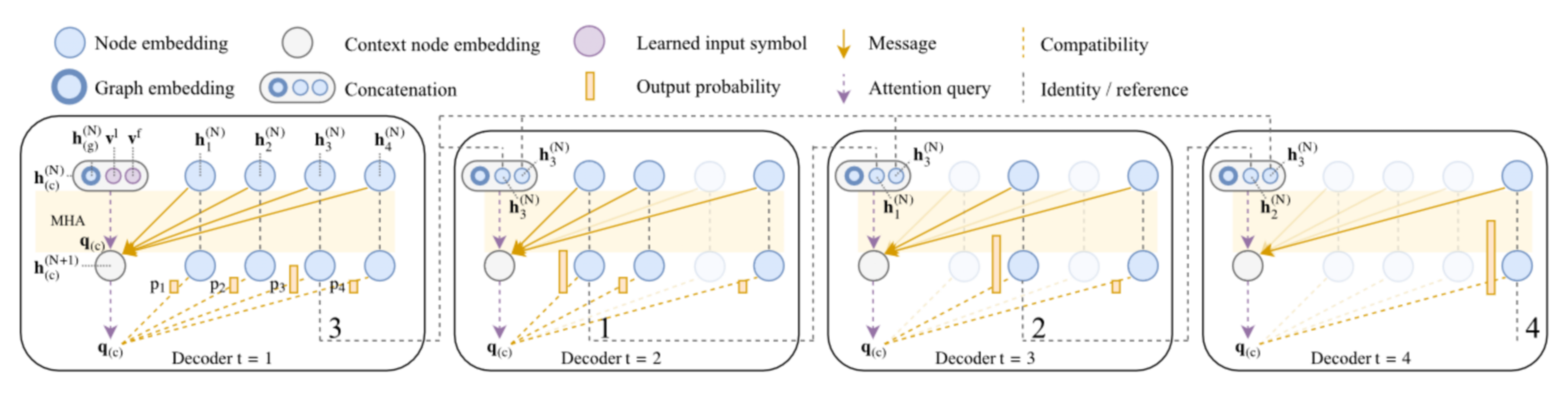}}
\caption{The decoder splices the encoder's output into a context embedding and calculates the selection probability by determining the attention between candidate points and the context embedding. Then, it selects them sequentially. (Credit to \cite{Kool2018AttentionLT})}
\label{fig:decoder}
\end{center}
\vskip -0.1in %-0.2in
\end{figure}

\section{Attention-based Decoder} \label{sec:decoder}
In this paper, we propose using the decoder introduced by \cite{Kool2018AttentionLT}, with its framework illustrated in the \cref{fig:decoder}. The encoder outputs the representation $h$ for each scenario and a global embedding $\overline{h}$. The decoder then selects scenarios sequentially with a multi-head attention mechanism. Kool et al. introduced a special context embedding $h_{(c)}$, created by horizontally concatenating the global embedding, the last selected scenario embedding, and the first selected scenario embedding. When $k=1$, with no scenario yet chosen, two learnable vectors serve as placeholders.

The decoder utilizes the context embedding as the query and the candidate points as the key and value to compute attention, applying a mask to the already selected ones. The output of the attention mechanism, after passing through the tanh, is clipped to the range [-10,10]. The final probabilities are then obtained using softmax on these values. This approach, focusing solely on attention calculations related to Context embedding, avoids an $n\times n$ attention computation (where $n$ is the number of candidates), significantly reducing computational complexity.

\section{Detail of Baselines} \label{sec:baselines}
\textbf{CSSC:} This method maps each instance into a space of cost values and clusters them in this space. It begins by obtaining optimal solutions for each scenario subproblem, then calculates the distance between scenario $i$ and $j$ by substituting the optimal solution of scenario $i$ into $j$ to get the value. Once all distances are determined, a clustering method is used to identify $K$ clusters and their respective representatives.

In practice, the CSSC method involves converting the cluster tasks into an equivalent Mixed Integer Programming (MIP) problem. This MIP is then solved using an advanced solver. To enhance efficiency, the solution of scenario subproblems is processed in parallel, utilizing two processes for the CSSC method.

In our experiments, we observed that although the CSSC method performs well, it requires excessive running time, particularly for instances with lots of scenarios.  For $N$ scenarios, it needs to solve $N$ subproblems and extend these solutions to other scenarios, involving $N^{2}-N$ computations. While manageable for 200 scenarios, the computational complexity becomes overwhelming as the number increases. Additionally, we find that the equivalent MIP problem of clustering often remains unsolved even after several hours, especially in many CFLP instances. Therefore, we limited our comparison of this method to the NDP$\_2\_2\_10\_200$. 

\textbf{NN-P and NN-E:} The two methods utilize ReLU neural networks to predict the objective value in the second stage. They then transform the network into a Mixed Integer Programming problem, which is accurately solved using a solver. For more information on the conversion technique, please see \cite{serra2018bounding, patel2022neur2sp}. 

In our experiments on CFLP and NDP datasets, we note that these methods require training and testing on the same instance, limiting their generalization and increasing time consumption. Therefore, we test in only 10 instances. For CFLP, we use the settings from \cite{patel2022neur2sp} for CFLP$\_10\_10$, and for NDP, the settings match those for CFLP$\_50\_50$.

However, the conversion methods used by NN-P and NN-E lead to the MIP problem whose size is linear to the total number of neural network nodes. Therefore, to enable quick solutions, the number of hidden layers and nodes in the neural network are kept relatively low. However, given the large number of scene parameters in our CFLP data (200 for CFLP$\_10\_20$) and the high number of decision variables in NDP (178 for NDP$\_2\_2\_10$), these networks with fewer nodes struggle to provide accurate predictions. Thus, due to these constraints, NN-P and NN-E are not very effective in handling our datasets.

\textbf{CVAE-SIP and CVAE-SIPA:} These methods use conditional variational autoencoders to learn the latent representation of scenarios, aiding in clustering and scenario reduction. In CFLP, facilities and customers are represented as vertices, connected by distance-weighted edges. For NDP, we follow the approach detailed in \cite{wu2022learning}. For CFLP,  due to the presence of the demand $q_{cf}^{s}$ of customer $c$ at facility $f$, we add $q_{cf}^{s}$ as a feature for the edge $(c,f)$.  For NDP, we use the pre-trained model provided, thanks to the consistent data. In CFLP, we adopt the training method described in the paper, utilizing 12,800 instances. For CVAE-SIPA, we used 1$\%$ of the instances to solve the objective values for all scenarios.

However, these methods are somewhat limited. They are specifically designed for graph-based stochastic planning problems, are challenging to generalize, and don't fully capture the intrinsic details of scenario sub-problems. They also don’t make complete use of solver feedback during training, suggesting room for further development

\section{Ablation Study}
The hierarchical graph network we propose involves extracting features from scenario bipartite subgraphs and integrating information in high-level instance graphs. To emphasize the significance of hierarchical graphs, we conduct ablation experiments by removing the high-level layers. We retain only the low-level graph network and directly feed its output into the decoder for scenario selection. Keeping training parameters constant, CFLP results are presented in the table, where our$\_$low indicates the results without high-level networks. Here, the performance significantly decreases, underperforming the original method at each $k$ and falling behind CVAE methods (except for CVAE-SIPA at $k=20$). This decline in performance highlights the crucial role of high-level information and the analysis of spatial relationships between scenarios. Additionally, the slight difference between our$\_$low and CVAE methods indicates the effectiveness of the low-level graphs. However, the model's performance is still constrained without considering the connections between scenarios.

\begin{table*}[t]
\caption{Ablation Study on CFLP$\_10\_20\_200$ error rates ($\%$) and time in seconds (s).}
\label{tab:cflp_10_20_200_ab_for_hign}
% \vskip 0.15in
\begin{center}
\begin{small}
\setlength\tabcolsep{1.5pt}
\begin{sc}
\begin{tabular}{lcccccc}
\toprule
 & \multicolumn{6}{c}{CFLP$\_10\_20\_200$}\\
 \hline
& \multicolumn{2}{c}{$k$=5} & \multicolumn{2}{c}{$k$=10} & \multicolumn{2}{c}{$k$=20} \\
Method & Error  & Time& Error & Time& Error & Time \\
\hline

CVAE-SIP& 6.60& 2.60 & 3.66 & 12.38 & 1.77 &\textbf{26.12} \\

CVAE-SIPA& 6.86&  2.40 & 3.49 & 10.23 & 2.50 &27.55 \\

HGCN2SP & \textbf{2.47($\pm$ 0.33)} & 2.45($\pm$ 0.37)  & \textbf{1.37 ($\pm$ 0.03)}& 14.72($\pm$ 1.35) & \textbf{1.16 ($\pm$ 0.01)} & 41.97($\pm$ 2.72)  \\

HGCN2SP$\_$low &  8.35 ($\pm$ 1.75) & \textbf{1.76($\pm$ 0.23)} &  4.34 & \textbf{9.34($\pm$1.74)} & 1.72($\pm$ 0.33) & 26.70($\pm$ 3.64) \\
\bottomrule
\end{tabular}
\end{sc}
\end{small}
\end{center}
\vskip -0.1in
\end{table*}

%%%%%%%%%%%%%%%%%%%%%%%%%%%%%%%%%%%%%%%%%%%%%%%%%%%%%%%%%%%%%%%%%%%%%%%%%%%%%%%
%%%%%%%%%%%%%%%%%%%%%%%%%%%%%%%%%%%%%%%%%%%%%%%%%%%%%%%%%%%%%%%%%%%%%%%%%%%%%%%

\end{document}